\documentclass[sigconf]{acmart}
\usepackage{graphicx}

\begin{document}

\title{Enhancing Human-robot Collaboration by Exploring Intuitive Augmented Reality Design Representations}


\author{Chrisantus Eze}
\orcid{0000-0001-5440-4316}
\affiliation{%
  \institution{Oklahoma State University}
  \city{Stillwater}
  \state{OK}
  \country{USA}}
\author{Christopher Crick}
\orcid{0000-0002-1635-823X}
\affiliation{%
  \institution{Oklahoma State University}
  \city{Stillwater}
  \state{OK}
  \country{USA}}



\copyrightyear{2023}
\acmYear{2023}
\setcopyright{acmlicensed}\acmConference[HRI '23 Companion]{Companion of the 2023 ACM/IEEE International Conference on Human-Robot Interaction}{March 13--16, 2023}{Stockholm, Sweden}
\acmBooktitle{Companion of the 2023 ACM/IEEE International Conference on Human-Robot Interaction (HRI '23 Companion), March 13--16, 2023, Stockholm, Sweden}
\acmPrice{15.00}
\acmDOI{10.1145/3568294.3580089}
\acmISBN{978-1-4503-9970-8/23/03}

\begin{abstract}
As the use of Augmented Reality (AR) to enhance interactions between
human agents and robotic systems in a work environment continues to
grow, robots must communicate their intents in informative yet
straightforward ways. This improves the human agent's feeling of trust
and safety in the work environment while also reducing task completion
time. To this end, we discuss a set of guidelines for the systematic
design of AR interfaces for Human-Robot Interaction (HRI)
systems. Furthermore, we develop design frameworks that would ride on
these guidelines and serve as a base for researchers seeking to
explore this direction further. We develop a series of designs for
visually representing the robot's planned path and reactions, which we
evaluate by conducting a user survey involving 14
participants. Subjects were given different design representations to
review and rate based on their intuitiveness and informativeness. The
collated results showed that our design representations significantly
improved the participants' ease of understanding the robot's intents
over the baselines for the robot's proposed navigation path, planned
arm trajectory, and reactions.
\end{abstract}



\maketitle

\section{Introduction}
There have been a burst of augmented reality (AR) applications to help
improve HRI in different fields, mainly in the industrial setting. As
we transition into an era of highly automated manufacturing, more
emphasis will be given to improving how human agents interact and
collaborate with robots in a work environment. These interactions have
yielded remarkable results so far, but many HRI challenges remain. One
such challenge is reducing the cognitive workload of a human agent
when interacting with a robot via an AR interface, especially one
unfamiliar with the system.  Poor communication of robot intention and
planned movements can cause hazards in the work environment that might
jeopardize safety, reduce task efficiency, and lead to flawed
perceptions of robot usability \cite{Walker2018}. However, active
research is currently being carried out to address these issues. AR
has had a significant impact on enhancing human-robot collaboration in
the workspace in different ways. It has been used for robot
programming and motion planning to instruct robots on task execution
\cite{GeorgiosRoman2021}, \cite{Hernandez2020}. This could be by
issuing either low-level \cite{GeorgiosRoman2021} or high-level goals
\cite{Hernandez2020} where the robot figures out the best way to
execute the request. AR has also been used to communicate the robot's
intentions, such as warning alerts, proposed path, arm motion, and
trajectory to the agent.

In our work, we outline and discuss a set of design guidelines that
should be followed when designing AR interfaces for HRI systems. These
guidelines not only ensure that interfaces are designed as intuitively
as possible for a novel user, but also provides systematic best
practices for an HRI researcher when developing HRI interfaces. In
order to illustrate the usage of the guidelines, we have developed
sets of interface designs for various types of robot feedback during
the course of a robot's interactions with a human agent in a work
environment. These interfaces include the proposed robot navigation path,
proposed robotic arm end-effector trajectory, and alert and warning
displays. Safety and trust of both the robot and the human agent(s)
are two important factors to be considered when designing HRI systems,
and in our work, we have explored the AR design space in order to
develop interfaces that would enhance these factors whilst also
ensuring a reduced task completion time. To evaluate our proposed
design frameworks, we created a survey where the participants gave
different ratings for the various interface designs.

\section{Related Work}
State-of-the-art manufacturing systems \cite{Hermann2016} has created
the need to reduce the complexity that arises from a hybrid work
environment \cite{ElMaraghy2005}. As noted in \cite{Lotsaris2021},
there should be tools that can conceal these complexities from human
agents and help them communicate and interact with the rest of the
system. Furthermore, \cite{Lotsaris2021} noted that these tools must
offer easy-to-use interfaces with an intuitive mechanism that should
not distract the workers from their tasks. Here, they
designed an application that doesn't require a human agent to have
specialized knowledge or training. However, the interface was
cluttered with many buttons, which could make the human agent have
difficulties figuring out the right button to press. In our work, we
tried to avoid this clutter by limiting the number of view items that
we overlay on the agent's screen.

Georgios et al. \cite{GeorgiosRoman2021} proposed a system that allows
for bidirectional communication between a robot and multiple users in
a work environment in real-time.  This work may not have represented
the design of their AR application in a way that would be intuitive
for the agents. For instance, in the paper, the end-effector's planned
trajectory was represented using a 3D sphere and the arm movement
radius as a red transparent sphere. However, using red for this
purpose could be misinterpreted to mean danger or a problem. Our
system explores different design representations for capturing the
robot's planned trajectory, making it easier for a human agent unfamiliar
with the system to grasp it quickly.

Lotsaris et al. \cite{Michalos2016} used an AR application to provide
functionalities such as robot motion and workspace visualization,
visual and audio alerts, and production data. Here, operators can
visualize all the components involved in the assembly sequence
intuitively using 3D models.  Their proposed approach, despite its
innovative audio alerts, is less informative than it could be.  The
text on the visual alert is blurry and the 3D model lacks sufficient
fidelity and might be challenging for a novel operator to understand
and navigate.  In contrast, in our proposed framework, we have ensured
that the design of the interfaces is very simple and easy even for an
untrained user. We also ensured that our 3D models have high fidelity.

Robot trajectory planning is one of the most important aspects of a
human-robot interaction system and research have been carried out to
improve and achieve better results. Fang et al. \cite{Fang2012}
designed a high-fidelity AR-based virtual manipulator which
facilitates robot programming and trajectory planning. Despite the
high fidelity of the interface and the accuracy achieved by the
simulated motion planner, the presented system would pose a challenge
to a less technical user given the level of technical sophistication
and low intuition provided by the system.


\section{AR Design Best Practices and Guidelines}
When designing AR interfaces for HRI purposes, it is important to
follow specific laid-out guidelines. These rules not only ensure that
interfaces are designed appropriately for new and experienced users,
but they also provide a systematic approach for the researcher towards
designing interfaces for HRI systems. Radkowski et al.
\cite{Radkowski2015} proposed that the level of difficulty of carrying
out a task during a manual assembly process should determine the
complexity of the user interface (UI). The factors to be considered
include the number of moving directions of a part, the number of
active surfaces the operator needs to keep aligned during a joining
procedure, the number of parts and operations in the assembly task,
the hierarchy of parts and operations in the assembly task, the
visibility of the part to assemble, and lastly, the posture of the
operator. However, the authors noted that these parameters can be
difficult to quantify, but with formal user studies using interface
prototyping, the assessment can be facilitated. Furthermore,
Danielsson \cite{Danielsson2016} discussed a concept called "design
science" which aims at developing ways to achieve human goals through
the creation of artifacts that are evaluated to ensure they meet the
design requirements. Wilkinson et al. \cite{Wilkinson2021} also
proposed some guidelines which should guide the design of interfaces
for HRI systems. The authors discussed the implementation of these
guidelines using two UIs they developed. We have also applied these
guidelines to our work.  Below is a description of the guidelines and
how we adhered to them.

\subsubsection{Understandable}
\cite{Wilkinson2021} stated that since HRI systems should be simple
even for an untrained user, the UIs must offer enough feedback and
intuition to allow for ease of understanding and use. Our proposed
the framework used basic visual features which most people are familiar
with, such as arrows and cones, thereby offering good intuition to the
users and reducing task completion time.

\subsubsection{Reliable}
Since the real world cannot be completely predictable, there are
chances that the user might make errors.  A well-designed HRI system
should have a robust error-handling capability
\cite{Wilkinson2021}. We adhered to this guideline by designing some
sets of visual alerts which give a user a proper visualization of the
status of the robot, enabling the user to avoid cases that could cause
harm to them or the robot.

\subsubsection{Accessible}
The HRI system should be able to be used by various people with
different physical and cognitive abilities. In our system, we ensured
that visual features such as labels, arrows, and cones are not only
accessible but also very legible and informative to all users.
Furthermore, \cite{Dnser2007} discusses some Human-Computer
Interaction (HCI) principles could serve as guidelines for designing
AR systems. We have extended and incorporated this set of guidelines
in our work. Below we discuss the guidelines most relevant to AR
applications for HRI purposes.

\subsubsection{Low Cognitive Overhead}
An AR interface should be designed in a way that would enable the user
to focus on the actual task and not get distracted by unimportant
information, helping reduce the agent's cognitive overhead. According
to \cite{Wilkinson2021}, registration errors in AR systems could
reduce the performance of users when the virtual elements from the AR
interface are not well aligned with the physical objects. The result
of this could lead to a reduction in task performance since
associating the virtual elements with the physical environment
requires cognitive effort from the user \cite{Dnser2007}.

\subsubsection{Learnable and Consistent}
A good user interface should enable users quickly to learn how to
navigate an AR application.  Maintaining a consistent interface, in
both appearance and functionality, is essential for this purpose. In
our work, we have ensured that we maintained a consistent look and
feel of the user interface. We used consistent color schemes, object
shapes, and dimensions to ensure uniformity and coherence.

\subsubsection{Responsive}
It is essential that AR applications are designed with responsiveness
in mind. It is almost impossible to rule out the occurrence of lags in
an application. Therefore, an AR application should be designed to
provide feedback to the user whenever such occurs. This helps keep the
user informed and less anxious.

The proposed design frameworks adhered to the above guidelines which
ensures a systematic approach to designing AR interfaces for HRI
systems in order to achieve informative and intuitive interfaces.

\begin{figure}[htbp] 
\includegraphics[scale=0.21]{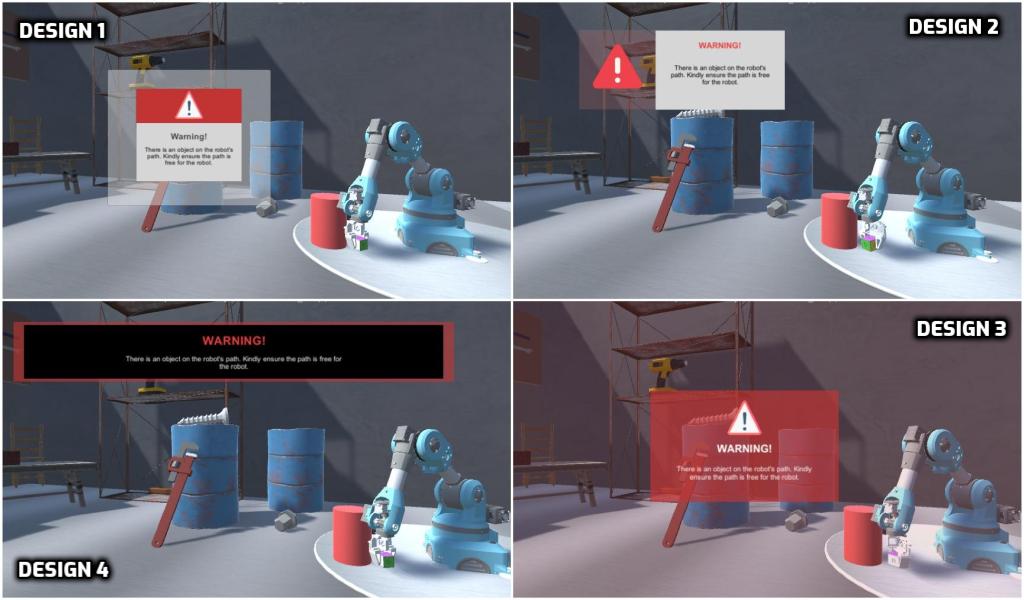} 
\caption{Different robot feedback warning alert designs.}
\Description{Different designs for representing robot reactions}
\label{fig:alert_collage}
\end{figure}

\section{Technical Description}
Our work uses a 3D/AR Unity application to explain and recommend the
different ways AR interfaces should be designed to enhance Human-Robot
collaboration in a work environment. We undertook an iterative process
to explore the design space of AR-HRI systems by providing intuitive
and concise feedback on robot motion intent and obstructions
identified by the robot. We evaluated several past AR-HRI systems
across different application scenarios and came up with a base for a
high-level framework on how to design and represent robot feedback in
an AR-HRI system for enhanced human-robot interactions.

We implemented our proposed HRI system using the open-source simulated
Niryo One robotic manipulator, a collaborative 6-axis robot in the
Unity 3D environment. It was provided by \cite{Niryo2022} in the form
of a Unified Robot Description File (URDF) \cite{UnityGithub2022},
which is similar to what was used in \cite{ShamaineMMSP2020}.

\begin{figure}[htbp] 
\includegraphics[scale=0.21]{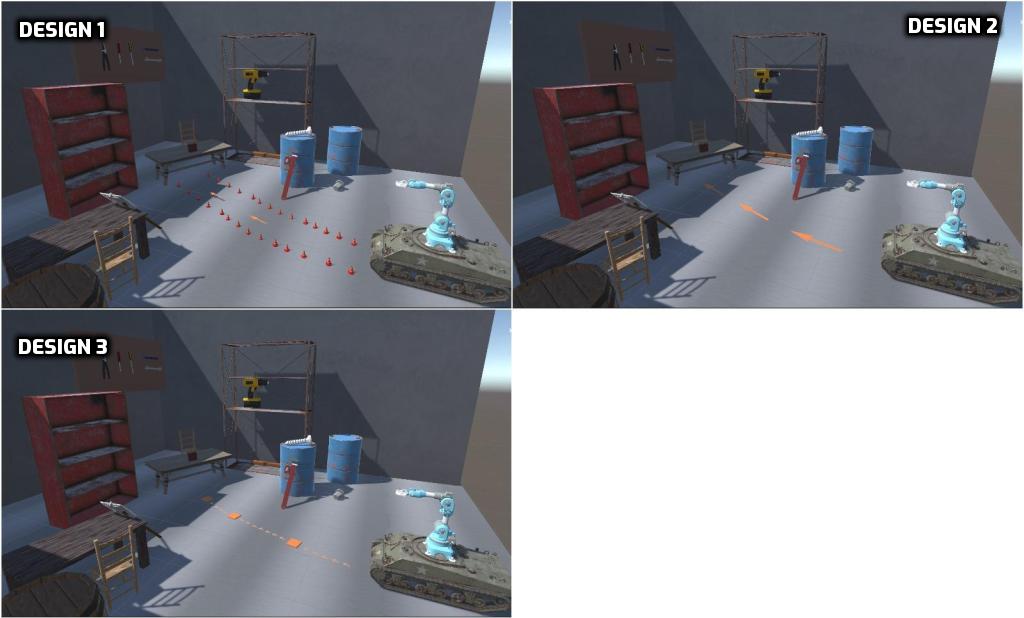} 
\caption{Different proposed robot navigation path designs}
\Description{Different designs for representing robot navigation paths}
\label{fig:path_collage}
\end{figure}

\section{Experimental Evaluation}
In this section, we will discuss the problems identified in the design
of interfaces for AR-HRI systems and evaluate our proposed
solutions. Our proposed solution tries to classify the interfaces into
three broad categories: visual alerts, robot navigation paths and
workspace visualization, and robotic arm end-effector manipulation and
object pickup. These designs were evaluated by recruiting 14 participants who
were presented with the different design representations of our
system in an online survey. The designs were presented in the order: path-arm-alert and they were asked to express their thoughts about the designs, how quickly they were able to figure out what the designs were about, and how intuitive the designs were. These responses in addition to the Likert scale ratings were collated from the participants.


\subsection{Visual Alerts}
Visual alerts could be general messages designated by the process
planner \cite{Makris2016} or more specific alerts like the path of the
robot. Due to the impact of these alerts in determining both user and
robot safety in the work environment, it is crucial to ensure that the
the information they convey is intuitive and easily understood by human
operators without requiring much cognitive effort.  The alert
interfaces in previous AR-HRI systems are often either illegible due
to poor choice of color and font or are cluttered with so many views
items that the messages are challenging for the human operator to
understand. This can be seen in \cite{Makris2016} and
\cite{Lotsaris2021}.

To address this challenge, we proposed four different visual
alert design representations (see Figure \ref{fig:alert_collage}) which were evaluated by the survey participants.

\begin{figure}[htbp] 
\includegraphics[scale=0.21]{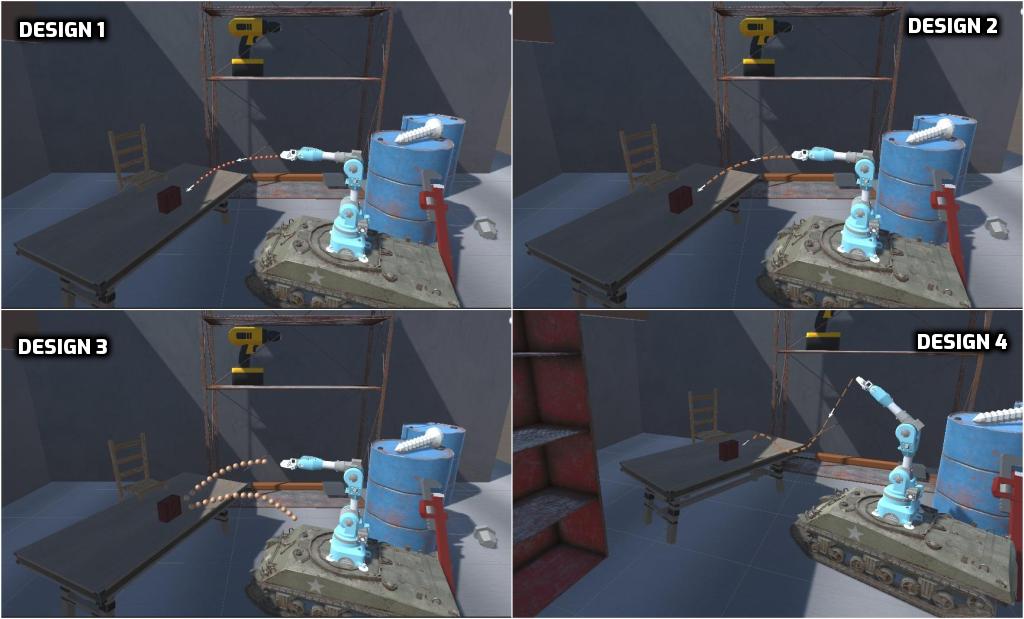} 
\caption{Different planned robot arm trajectory designs}
\Description{Different designs for representing robot arm trajectory and motion}
\label{fig:arm_collage}
\end{figure}

\begin{figure*}[htbp] 
\includegraphics[scale=0.25]{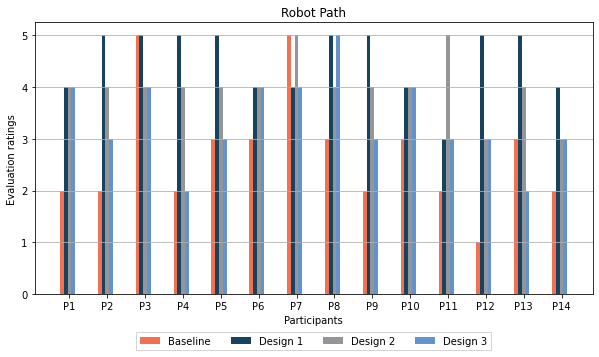} 
\includegraphics[scale=0.25]{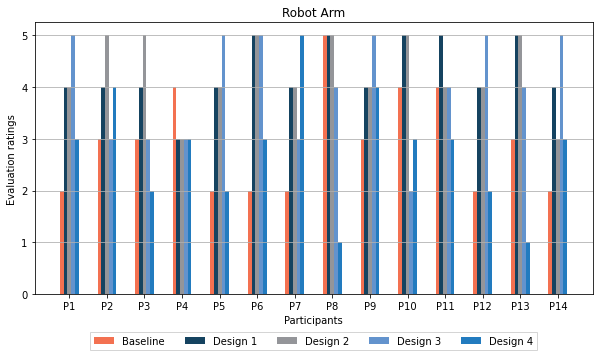} 
\includegraphics[scale=0.25]{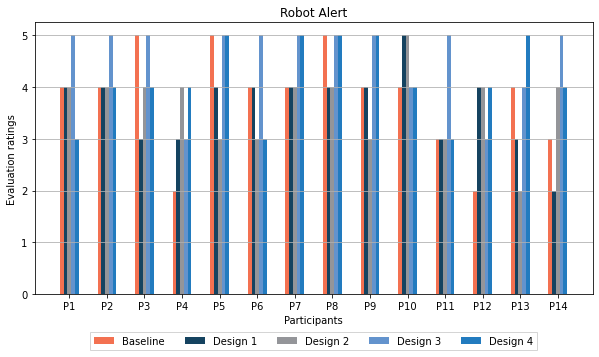} 
\caption{User ratings for the robot proposed navigation path, planned arm trajectory, and alert designs}
\Description{This shows the result from the online survey carried out}
\label{fig:results}
\end{figure*}

\subsection{Robot Navigation Path and Workspace Visualisation}
Robot navigation and path planning are essential aspects of an HRI
system. A poorly designed visual representation of the robot's planned
the path is likely to lead to confusion in the work environment among the
human operators, which not only has the potential to harm the
operators but also increases the production time. In
\cite{GeorgiosRoman2021}, the planned robotic path was represented
using a set of 3D green spheres and has no similarity with real-world
path indicators, introducing the potential of being
misunderstood by an operator unfamiliar with the system and
environment.

We have addressed this challenge by applying familiar real-world
design representations and concepts such as traffic cones and road
barriers to developing three different design representations
of robot navigation paths (see Figure \ref{fig:path_collage}) that have been evaluated and rated by the survey participants.

\subsection{Robotic Arm End-Effector Manipulation and Object Pickup }
Augmented reality has often been used to plan the path and orientation
of the end-effector of a robot. However, adequate measures should be
taken to ensure that these paths are represented in the design space
intuitively, avoiding complexity whenever possible.  Just like their
proposed robot navigation path, \cite{GeorgiosRoman2021} represented
the proposed path of their end-effector using a set of green 3D
spheres that made an oval shape. This design concept could potentially
confuse the operators, especially those unfamiliar with the system
because they failed to use familiar and conventional view elements to
represent paths and trajectories in the physical world.  To address
this, we have evaluated several design approaches to represent the
planned paths and trajectories of end-effectors (see Figure
\ref{fig:arm_collage}).

\section{Results \& Discussions}
This section summarises our findings based on the survey involving 14
participants. The designs were reviewed based on their intuitiveness,
informativeness, and proximity to natural and familiar
representations. The participants were asked to rate the proposed
methods and the baseline methods (\cite{GeorgiosRoman2021}, \cite{Michalos2016}, and \cite{Fang2012}) on a 5-point Likert scale (see Figure \ref{fig:results}).

\subsection{Proposed Navigation Path}
The first subfigure shows the survey results for the different robot
navigation path designs.  Most of the participants had difficulty
understanding the robot's intent when they were presented with the
baseline's representations. However, our approach significantly
improved the participants' understanding of the robot's intent.  The
Design 1 representation was the most informative and imposed the
least cognitive overload to the participants.

\subsection{Planned Arm Trajectory}
Results for the proposed robotic arm trajectory designs are
shown in the second subfigure. In this category, the baseline
approaches were also rated as less informative by most of the
participants compared to our proposed approaches.  However, most
participants rated their understanding of the proposed designs as
either neutral or informative. Thus, the proposed approaches did not
significantly improve the users' understanding of the robot's intent.

\subsection{Alert}
The third subfigure shows the survey results for the various robot reactions designs, such
as warning alerts. The participants rated our baseline approaches as informative. In contrast, our proposed designs
provided a more explicit representation of the robot's intent to the
participants and most participants rated them as either informative or
very informative.

We found a significant effect of design on perceived
communication clarity, which translates to the user's perception of
the robot as a good work partner. Aside from the ratings of the
presented designs, some participants left important feedback regarding
how they felt during their evaluations of the different designs.

\begin{itemize}
\item Participant X [Arm Trajectory]: "At first, I thought
  the robot was shooting cannon balls at the object."
\item Participant Y [Arm Trajectory]: "Is this trying to
  represent projectile motion or something of that sort?"
\item Participant Z [Navigation Path]: "I initially didn't
  understand what the dotted colored lines on the floor meant. It
  took me a while to figure it out."
\end{itemize}

Interestingly, some participants found some of the design representations from
the baseline methods to be more informative than some of our
design representations in some categories. However, the same
participants rated at least one of our proposed designs higher than
all the baseline approaches, even in those cases.

\section{Conclusions \& Recommendations}
This research explored the design space of human interactions with a
robot through AR to convey the robot's intentions. We conducted a
survey involving 14 participants, where each participant was presented
with different design representations which they rated against the
baseline approaches from previous work. We found that they were
able to easily and more quickly understand the intents of the robot at
the different stages of the human-robot collaboration cycle. From the results, we note that our proposed representations significantly improved their understanding of the robot's
intents over the baseline approaches.

The aim of this work was to introduce a systematic approach to
designing AR interfaces for HRI systems to ensure that they adhere to
the best practice of providing a very limited cognitive workload to
the users. From the statistics of our collated results, we could see
the impact our proposed design representations made on the
participants in terms of the ease of understanding the robot's
intents, hence we believe this aim was achieved. Therefore, we
recommend our approach as a base for researchers exploring ways to
improve the usability of their HRI systems, especially when designing
their robot's visual feedback systems for conveying robot intentions
and messages to human agents.


\bibliographystyle{ACM-Reference-Format}
\bibliography{bibliography}

\end{document}